\documentclass[letterpaper, 10 pt, conference]{ieeeconf}  

\IEEEoverridecommandlockouts                              

\usepackage{todonotes}
\usepackage{graphicx}
\usepackage{multirow}
\usepackage[acronym,toc,nomain]{glossaries}

\makeglossaries

\newacronym{cnn}{CNN}{Convolutional Neural Network}

\newacronym{lstm}{LSTM}{Long Short-Term Memory}

\newacronym{gmm}{GMM}{Gaussian Mixture Model}

\newacronym{svm}{SVM}{Support Vector Machine}

\newacronym{fdd}{FDD}{Fall Detection Database}

\newacronym{roc}{ROC}{Receiver Operating Characteristic}

\newacronym{tp}{TP}{True Positive}

\newacronym{tn}{TN}{True Negative}

\newacronym{fp}{FP}{False Positive}

\newacronym{fn}{FN}{False Negative}

\newacronym{fc}{FC}{Fully-Connected}

\title{\LARGE \bf
Fall Detector Adapted to Nursing Home Needs through an Optical-Flow based CNN
}

\author{Alexy CARLIER$^{1}$$^{*}$, Paul PEYRAMAURE$^{1}$$^{*}$, Ketty FAVRE$^{2}$ and Muriel PRESSIGOUT$^{1}$
\thanks{$^{1}$Univ Rennes, INSA Rennes, CNRS, IETR - UMR 6164, Rennes, France.}%
\thanks{$^{2}$Univ Rennes, CNRS, IETR - UMR 6164, Rennes, France.}%
\thanks{$^{*}$Both authors contributed equally to this work.}%
}

\begin{document}

\maketitle
\thispagestyle{empty}
\pagestyle{empty}

\begin{abstract}

Fall detection in specialized homes for the elderly is challenging. Vision-based fall detection solutions have a significant advantage over sensor-based ones as they do not instrument the resident who can suffer from mental diseases. This work is part of a project intended to deploy fall detection solutions in nursing homes. The proposed solution, based on Deep Learning, is built on a \gls{cnn} trained to maximize a sensitivity-based metric. This work presents the requirements from the medical side and how it impacts the tuning of a \gls{cnn}. Results highlight the importance of the temporal aspect of a fall. Therefore, a custom metric adapted to this use case and an implementation of a decision-making process are proposed in order to best meet the medical teams requirements.
\newline

\indent \textit{Clinical relevance}— This work presents a fall detection solution enabled to detect 86.2\% of falls while producing only 11.6\% of false alarms in average on the considered databases.
\newline

\end{abstract}

\vspace*{-1em}
\section{INTRODUCTION}
\label{sec:intro}

\glsresetall

In specialized homes for the elderly, fall is the leading cause of death due to trauma as a resident falls 1.7 times a year in average in France~\cite{podvin-deleplanque_fall_2015}. Some of them being more or less prone to falls, medical teams may discover a person who has fallen to the ground only after several hours~\cite{fleming_inability_2008}. In this context, a fall detector must detect falls while avoiding false alarms unnecessarily disturbing to the medical staff, which can not afford too frequent and intuitive interruptions. According to a study~\cite{fraudet_silver_2020} conducted with specialized medical teams, residents and families in three different nursing homes, the solution must:

\begin{itemize}
	\item detect as many falls as possible;
	\item give no false alarms;
	\item not be an extra equipment to be worn by the resident;
	\item be re-configurable and adaptable to different residents.
\end{itemize}

Fall detection solutions are divided into two types of approaches: sensor-based and vision-based. This work focuses on vision-based solutions since wearable sensor-based ones do not meet medical staff requirements. Indeed, they are not adequate when dealing with people suffering from mental diseases which is more frequent with older people. Moreover, even if a camera can cause privacy issues, the study shows that majority of medical teams, residents and families approve its use for residents safety and independence. 

In general, a fall leads to a change of the human body velocity and position. Thus, in image-based techniques, features such as 2D human body pose estimation~\cite{lie_human_2018}, movement vectors or person silhouettes using background subtraction algorithms~\cite{debard_camera-based_2015} are extracted from images. These features, enabling to locate the person and know its spatial body orientation, are usually fed to a classifier such as \gls{gmm}~\cite{rougier_robust_2011} or \gls{svm}~\cite{charfi_definition_2012}~\cite{zerrouki_fall_2016}. The difficulty to characterize falls led the community to consider its temporal aspect. The use of \gls{lstm}~\cite{musci_online_2018} networks enables advances in vision-based fall detection. Another approach is to feed several RGB images directly to a 3D \gls{cnn}~\cite{li_pre-impact_2019}~\cite{solbach_vision-based_2017} to exploit their temporal aspect.

Due to disabilities and/or old age, elderly falls can be characterized by different types and velocities of movement: hard falls (from standing position) and soft falls (from another starting point). They are therefore very difficult to model. Thus, a solution based on a neural network that takes into account human body motion changes seems to be adapted to this problem. The presented solution is optical-flow based and uses a \gls{cnn} originally trained to maximize a sensitivity metric~\cite{nunez-marcos_vision-based_2017}. We propose a different training approach using realistic metrics adapted to the application as well as a decision-making process adjustment which minimizes the number of false alarms and ensures a sufficient correct detection rate according to medical staff requirements.

\section{METHODOLOGY}
\label{sec:pagestyle}

\subsection{Solution overview}
\label{subsec:solution_overview}

\begin{figure}[!htb]
	\vspace*{0.3em}
	\centering
	\includegraphics[scale = 0.32]{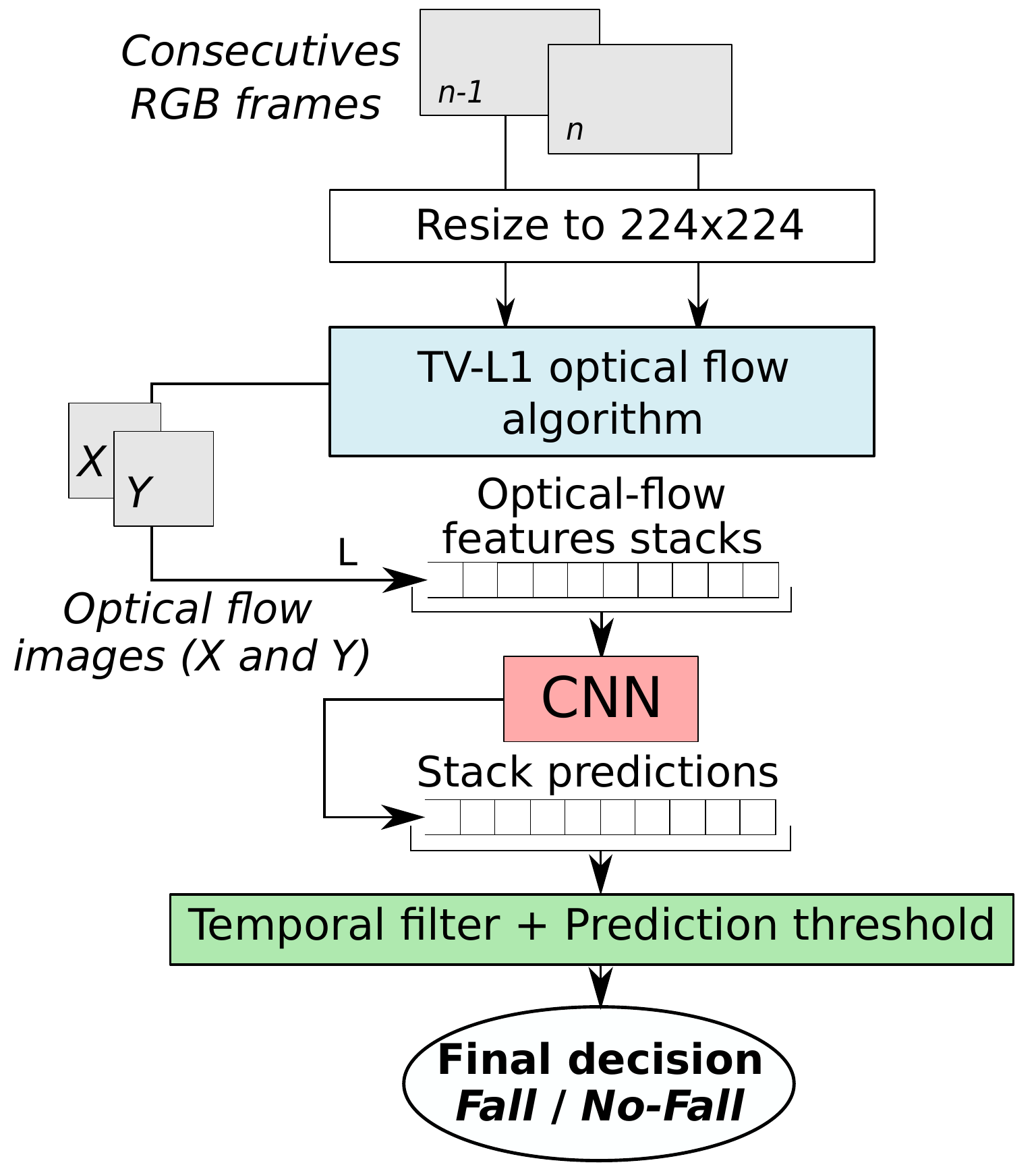}
	\vspace*{-0.7em}
	\caption{Solution overview from the input (RGB frames) to the output (final decision \textit{Fall} / \textit{No-Fall})}
	\label{fig:solution_overview}
	\vspace*{-1.3em}
\end{figure}

The general solution presented in Fig.~\ref{fig:solution_overview} is made of three stages. The first step takes as input two consecutive RGB images from a camera to generate a pair of optical flow images using the dense optical-flow TV-L1 algorithm~\cite{sanchez_tv-l1_2013}. The second step of the solution is a custom VGG-16 \gls{cnn} designed as in~\cite{wang_towards_2015} and pre-trained as in~\cite{nunez-marcos_vision-based_2017}. It takes as input a stack $S$ of $L=10$ consecutive pairs of optical flow images and infers a fall prediction. Finally, a custom temporal filter and a prediction threshold are applied to the \gls{cnn} output in order to exploit the fall temporal aspect. In this way, a single or consecutive stacks $S$, labeled as \textit{Fall} at the temporal filter output, raise a fall alarm to the medical staff.

\subsection{Databases and training procedure}
\label{subsec:databases&training_methodology}

Three labeled fall video databases (URFD~\cite{kwolek_human_2014}, FDD~\cite{noauthor_fall_nodate}, Multicam~\cite{auvinet_multiple_nodate}) are used for the training and test processes. They are composed of videos containing a fall action or not. Those which contain daily-life actions are annotated as \textit{non-fall}. Fall videos are split into three parts: the \textit{pre-fall} which is the action time before the fall, the \textit{fall} action, and the \textit{post-fall} which often corresponds to a person lying on the ground. The classifier of the \gls{cnn} has two output classes: \textit{Fall} and \textit{No-Fall}. \textit{Fall} class contains \textit{fall} actions and \textit{No-Fall} class takes into account \textit{pre-fall}, \textit{post-fall} and \textit{non-fall} video sequences. Fall action databases usually contain more daily-life actions than falls. In order to fine-tune the classifier, which is composed by two last \gls{fc} layers of the \gls{cnn}, and overcome unbalanced data, a weighted binary cross-entropy loss function (\ref{eq:loss}) is adopted as proposed in~\cite{nunez-marcos_vision-based_2017}. In that equation, $p$ is the prediction of the network, $t$ is the ground-truth, the class weight $w_0$ is associated to \textit{Fall} class and $w_1$ to \textit{No-Fall} class.
\begin{equation}\label{eq:loss}
loss(p,t)=-(w_1~.~t~.~log(p)+w_0~.~(1-t)~.~log(1-p)) 
\end{equation}

Leveraging on transfer-learning, it is possible to achieve fall detection while the amount of data is limited. All network layers except the last two \gls{fc} layers are frozen. That two last \gls{fc} are trained with fall databases using the 5-cross fold validation~\cite{refaeilzadeh_cross-validation_2009}. In order to have a training framework usable for any other database, our training approach differs from~\cite{nunez-marcos_vision-based_2017} on these points:
\begin{itemize}
	\item During the 5-cross fold validation, each video sequence is entirely stored in a single fold. It avoids having similar stacks in the train set and in the test set. Moreover, the train set and the validation set are filled with different videos in order to avoid overfitting. Then for a given initial fall video sequence, the derived sequences \textit{pre-fall}, \textit{fall} and \textit{post-fall} are in the same fold.
	\item At testing time, a \textit{Transition} class is used in addition to \textit{Fall} and \textit{No-Fall} classes in order to bring realistic cases. This class contains frames at the transition between \textit{pre-fall} and \textit{fall} and the transition between \textit{fall} and \textit{post-fall} sequences. These particular testing frames are not used in~\cite{nunez-marcos_vision-based_2017} and~\cite{wang_towards_2015}.
\end{itemize}

In order to obtain the best training efficiency possible, a grid search methodology is used with the following hyper-parameter ranges:
\begin{itemize}
	\item Learning rate $\lambda$: \{10$^{-2}$, 10$^{-3}$, 10$^{-4}$, 10$^{-5}$, 10$^{-6}$\}
	\item Batch size $Bs$: \{128, 256, 512, 1024\}
	\item Class weight $w_0$: \{1, 2, 5, 10, 15, 20\} and $w_1$: \{1\}
	\item Classifier activation function $f_{act}$: \{ELU, ReLU\}
\end{itemize}

Configurations are evaluated through specificity $sp$ (\ref{eq:metric_sp}), sensitivity $se$ (\ref{eq:metric_se}) and precision $p$ (\ref{eq:metric_p}) where $TP$ stands for True Positives, $TN$ for True Negatives, $FP$ for False Positives and $FN$ for False Negatives. These metrics are computed over stack predictions and are used to choose the best hyper-parameters configuration.
\begin{equation}\label{eq:metric_sp}
\begin{aligned}
sp={TN}/(TN+FP)
\end{aligned}
\end{equation}
\begin{equation}\label{eq:metric_se}
\begin{aligned}
se={TP}/(TP+FN)
\end{aligned}
\end{equation}
\begin{equation}\label{eq:metric_p}
\begin{aligned}
p={TP}/(TP+FP)
\end{aligned}
\end{equation}

In~\cite{nunez-marcos_vision-based_2017}, authors focus on maximizing sensitivity which leads to a decrease of specificity and precision. In our case, according to medical staff requirements which are explained in section~\ref{sec:intro}, specificity and precision are favored and it must be a trade-off with the sensitivity.

\subsection{Alarm precision oriented fall evaluation}
\label{subsec:Precision_oriented_fall_evaluation}

It is difficult to determine the exact beginning and end of a fall which makes it a complex event to characterize. Furthermore, the duration of a fall must be taken into account when evaluating the predictions. As presented in section~\ref{subsec:solution_overview}, the network makes a prediction with $L$ consecutive optical flow images. In the considered databases, the average fall duration is 1.11 seconds as seen in Table~\ref{tab:datasets_properties}. This means that a fall prediction is made during 1/3 of the average fall duration considering a 30 FPS recording. The addition of a temporal filter, as defined below, reinforces the time aspect of a fall and aggregates safe signals while reducing false alarms.

\vspace*{-0.5em}
\begin{table}[!htb]
	\caption{Databases properties}
	\vspace*{-1.5em}
	\begin{center}
		\begin{tabular}{ccccc}
			\hline
			\multirow{2}{*}{Database} & Frame rate & Avg. fall duration & Number\\ 
			& (FPS) & (frames - seconds) & of falls\\ 
			\hline
			URFD & 30 & 30 - 1.00 & 30\\ 
			FDD & 25 & 24 - 0.96 & 99\\ 
			Multicam & 30 & 41 - 1.36 & 200\\ 
			Avg. & 28 & 32 - 1.11 & -\\ \hline
		\end{tabular}
	\end{center}
	\label{tab:datasets_properties}
\end{table}
\vspace*{-0.5em}

In the temporal analysis step, predictions are considered no longer as stacks but as consecutive identical stack prediction types. They are labeled either as $TP_{a}$ for a true fall alarm, as $FP_{a}$ for a false fall alarm, or as $FN_{a}$ for a miss-detected fall. The implemented convolution filter is modeled by a gate function. It is defined by its width $W$ (in frames or seconds) conjointly tuned with the prediction threshold $T_{pred}$ of the filter. Below $T_{pred}$, a prediction is labeled as fall. These parameters are adjusted with the aim of minimizing the number of false alarms without missing falls. To measure this capability, $F_\beta$ (\ref{eq:F_beta}) is a function of the alarm precision $p_{a}$ (\ref{eq:p_{a}}) and the alarm sensitivity $se_{a}$ (\ref{eq:se_{a}}) in the same spirit as in~\cite{debard_camera-based_2016}. When $0<\beta<1$, $F_\beta$ metric weighs sensitivity less than precision by emphasizing more on false alarms and inversely when $\beta>1$. This metric enables a realistic fall detector evaluation with respect to the medical requirements.
\begin{equation}\label{eq:p_{a}}
\begin{aligned}
p_{a} = {TP_a}/(TP_a+FP_a)
\end{aligned}
\end{equation}
\begin{equation}\label{eq:se_{a}}
\begin{aligned}
se_{a} = {TP_a}/(TP_a+FN_a)
\end{aligned}
\end{equation}
\begin{equation}\label{eq:F_beta}
\begin{aligned}
F_\beta = (1+\beta^{2})~.~\dfrac{p_a~.~se_a}{(\beta^{2}~.~p_a) + se_a}
\end{aligned}
\end{equation}

\section{EXPERIMENTS}
\label{sec:typestyle}

\subsection{Hyper-parameters choice}
\label{subsec:hyperparameter_choice}

Hyper-parameters of the \gls{cnn} training are adjusted regarding the previously exposed specifications, namely a trade-off between high specificity and sufficient sensitivity.

The first hyper-parameter to be tuned is the \textbf{learning rate} $\lambda$. From the studied values, $ 10^{-2} $ is too high and causes the model to diverge. On the other hand, the model converges too slowly for a learning rate lower than $ 10^{-4} $. 

Concerning the \textbf{batch size} $Bs$, the choice made in~\cite{nunez-marcos_vision-based_2017} (i.e. 1024) may not lead to a well converged model as it was too high. From our experiments, a smaller batch size (of 128 or 256) leads to a better specificity. It deteriorates the sensitivity $se$ due to an increase of $FN$ but leads to a small impact on the alarm sensitivity $se_a$. 

The \textbf{activation function} ELU leads to a better sensitivity than the ReLU activation function that gives a better specificity. ReLUs are therefore chosen for our use case. 

Finally, a \gls{roc} analysis is made on the \textbf{class weight $w_0$} ($w_1$ is arbitrarily set to 1) to put emphasis on \textit{Fall} class and select the configuration giving the best specificity. In practice, $w_0$ higher than 5 implies instabilities in results both with balanced and unbalanced amounts of data in each class. $w_0$ set to 2, as in~\cite{nunez-marcos_vision-based_2017}, slightly increases the specificity and avoids overfitting on the $Fall$ class. Best configurations giving a well-trained model with an acceptable specificity are summarized in Table~\ref{fig:training_results}.

\vspace*{-0.5em}
\begin{table}[htb]	
	\caption{Best training configurations and associated results (in \%)}
	\centering
	\vspace*{-0.5em}
	\begin{tabular}{ccccc|cccccc}
		\hline  
		\multirow{2}{*}{ID}&\multirow{2}{*}{$\lambda$}&\multirow{2}{*}{$w_0$}&\multirow{2}{*}{$Bs$}&\multirow{2}{*}{$f_{act}$}&\multicolumn{2}{c}{URFD} & \multicolumn{2}{c}{FDD} & \multicolumn{ 2}{c}{Multicam}\\ 
		 \cline{6-11}
		 &  & & & & $se$ & $sp$ & $se$ & $sp$ & $se$ & $sp$\\ \hline
		1 & $10^{-3}$ & 2 & 128 & ReLU & 95.5 & 93.2 & 94.7 & 97.5 & \textbf{56.5} & \textbf{99.4}\\
		2 & $10^{-3}$ & 2 & 256 & ReLU & \textbf{89.5} & \textbf{94.1} & \textbf{93.5} & \textbf{97.6} & 59.4 & 99.0 \\ 
		3 & $10^{-4}$ & 2 & 128 & ReLU & 93.5 & 88.7 & 95.2 & 97.5 & 68.2 & 96.0 \\
		4 & $10^{-4}$ & 2 & 256 & ReLU & 95.5 & 89.2 & 96.0 & 96.9 & 71.4 & 93.7\\ \hline
	\end{tabular}
	\label{fig:training_results}
	\vspace*{-1em}
\end{table}

\subsection{Temporal Analysis}
\label{subsec:data_analyzis}

\begin{figure}[htb]
	\centering
	\includegraphics[scale = 0.43]{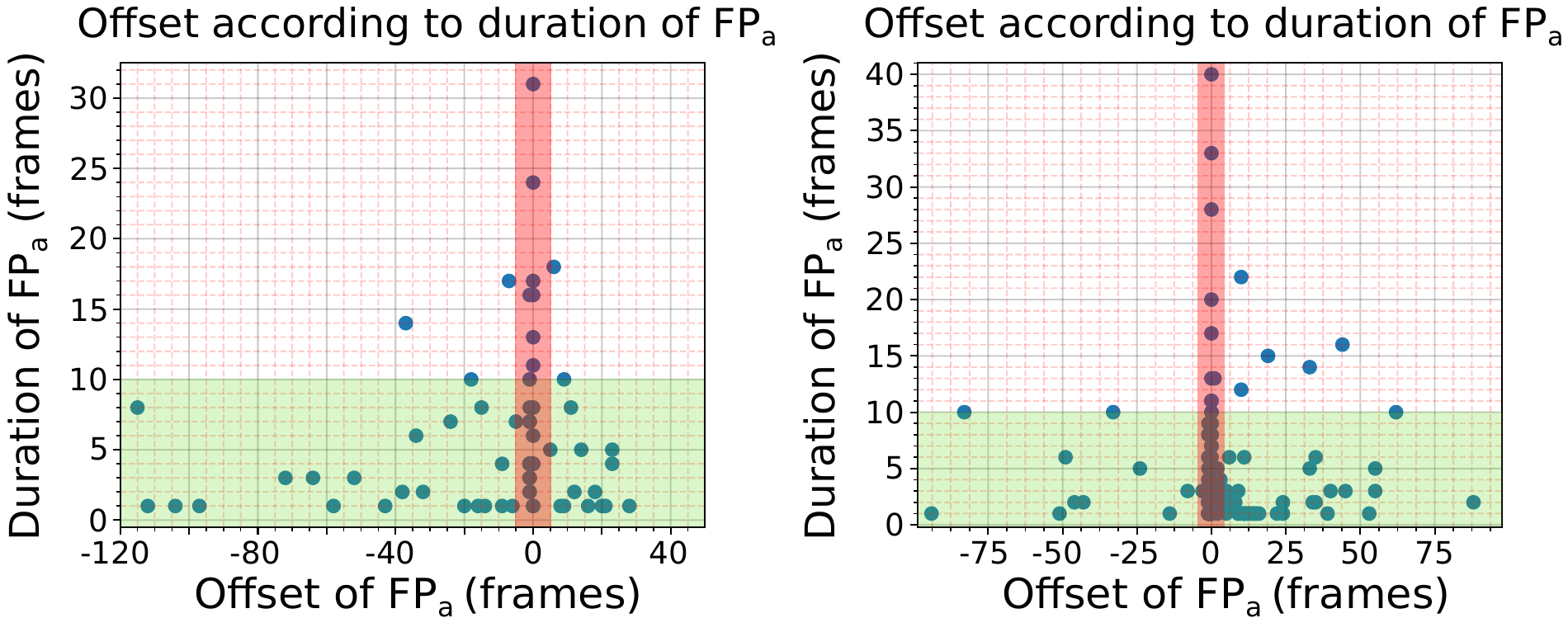}
	\caption{Offset to the \textit{Fall} class according to the duration of $FP_{a}$. The horizontal zone (in green) includes $FP_{a}$ duration shorter than 10 frames. The vertical zone (in red) includes $FP_{a}$ offset smaller than 5 frames. URFD on the left and FDD on the right.}
	\label{fig:offset}
\end{figure}

In order to improve model performances, prediction results are analyzed during training following their temporal aspect using metrics presented in section~\ref{subsec:Precision_oriented_fall_evaluation}. An analysis of the number of frames between $FP_{a}$ and \textit{Fall} labels, defined as offset in Fig.~\ref{fig:offset}, allows to better characterize false alarms.

In the three databases, $39\%$ of $FP_{a}$ are very close in time to the actual fall with an offset smaller than 5 frames. These predictions are labeled as false according to the ground-truth but are ambiguous. Indeed, they could be considered as the beginning or the end of the related fall from a human perception. Secondly, $86\%$ of $FP_{a}$ are shorter than 10 frames and will be removed by the temporal filter application.

Table~\ref{table:temporal_results_initial} shows the evaluation results of the \gls{cnn} output predictions, with $T_{pred} = 0.5$, using the theoretical precision metric $p$ computed with stack predictions, the alarm precision $p_{a}$ and the alarm sensitivity $se_{a}$. Concerning all databases, $p_{a}$ is significantly lower than $p$ which is expected as the temporal property is not taken into account yet.
\vspace*{-0.5em}
\begin{table}[htbp]
	\caption{Temporal evaluation of the CNN output predictions with $T_{pred} = 0.5$ and training configuration ID 2 (in \%) }
	\vspace*{-1.5em}
	\begin{center}
		\begin{tabular}{cccc}
			\hline
			Database & $p$ & $ p_{a} $ & $se_{a}$ \\ \hline
			URFD & 47.0 & 27.0 & 100 \\ 
			FDD & 59.1 & 54.7 & 98.9 \\ 
			Multicam & 63.6 & 25.5 & 89.0 \\ 
			Avg. & 56.6 & 35.7 & 96.0 \\ \hline 
		\end{tabular}
	\end{center}
	\label{table:temporal_results_initial}
	\vspace*{-1.5em}
\end{table}

\subsection{Filter size and prediction threshold adjustments}
\label{subsec:filter_implementation}

\begin{figure}[htbp]
	\centering
	\includegraphics[scale = 0.43]{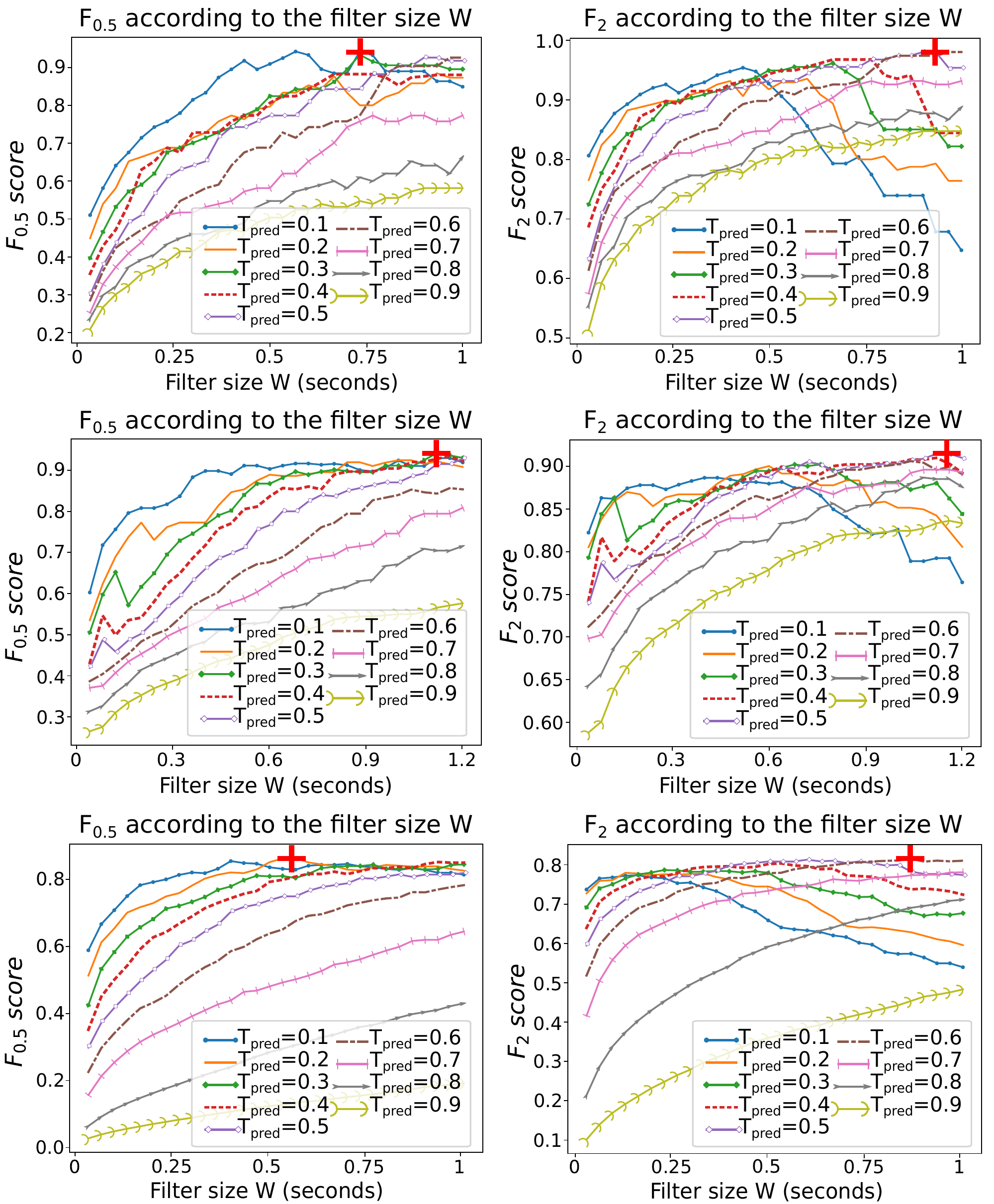}
	\caption{$F_{0.5}$ (at left) and $F_{2}$ (at right) according to filter size $W$ in seconds. Each graph plots curves for a $T_{pred}$ ranging from 0.1 to 0.9. Red crosses annotate configurations reaching the maximal $F_\beta$. From top to bottom, graphs correspond to the three databases: URFD, FDD and Multicam.}
	\label{fig:filter_temporal_study}
\end{figure}

An empirical study, illustrated in Fig.~\ref{fig:filter_temporal_study}, is made in order to propose the best association between the temporal filter size $W$ and the prediction threshold $T_{pred}$. Our use case suggests to maximize the alarm precision $p_{a}$ while stabilizing the alarm sensitivity $se_{a}$ on the three considered databases. 

$F_{0.5}$ and $F_{2}$ decrease as $T_{pred}$ is getting larger due to the rise of undetected falls. From a certain $W$ value, $F_{2}$ is reduced due to the same reason. For Multicam database, the number of undetected fall sequences is important and comes from the fact that it is a complex database with inaccuracies in labeling. The sequences are also much longer and more complex in terms of action. The objective is to drastically decrease the number of false alarms while limiting the number of undetected falls. In our case, $p_{a}$ is fixed to be more than 80\% and $se_{a}$ must not vary more than 10\% from Table~\ref{table:temporal_results_initial}. The optimal ($W$, $T_{pred}$) pair is found by averaging ($W$, $T_{pred}$) pairs that maximize $F_\beta$ for each database. 

In the end, $W = 0.87$ sec and $T_{pred} = 0.4$ are found to be the best combination and lead to the results in Table~\ref{table:final_results}. Compared to Table~\ref{table:temporal_results_initial}, the alarm precision $p_{a}$ increases drastically from 35.7\% to 88.4\% with the optimal ($W$, $T_{pred}$). On the other hand, the alarm sensitivity $se_{a}$ decreases from 96.0\% to 86.2\% per database in average. This means that the solution detects 86.2\% of falls while 88.4\% of the raised alarms are real falls. 
Considering the best camera (on which the fall is best visible) on Multicam database, the results of our solution are 7\% higher in terms of alarm precision and 4\% in terms of alarm sensitivity than in~\cite{debard_camera-based_2016}.

\vspace*{-0.5em}
\begin{table}[htbp]
	\caption{Final results (in \%) with the tuned decision process: \newline $W$ = 0.87 sec and $ T_{pred}$ = 0.4}
	\vspace*{-1.5em}
	\begin{center}
		\begin{tabular}{cccccccc}
			\hline
			Database &$F_{0.5}$&$F_{2}$ & $ p_{a} $ & $se_{a}$&$TP_a$&$FP_a$ & $FN_a$ \\ \hline
			URFD &87.4&94.2& 85.3 & 96.7 & 29 & 5 & 1 \\ 
			FDD &92.4& 91.3 & 92.8 & 90.9 & 90 & 7 & 9 \\ 
			Multicam & 83.3 &73.7& 87.1 & 71.0 & 142 & 21 & 58 \\ 
			Avg. & 87.7 & 86.4 & 88.4 & 86.2 & - & - & - \\ \hline
			Multicam$^1$ & \textbf{89.1} & \textbf{91.3} & \textbf{88.5} & \textbf{92.0} & \textbf{23} & \textbf{3} & \textbf{2} \\ 
			Multicam$^2$ & 82.7 & 86.6 & 81.5 & 88.0 & 22 & 5 & 3 \\ \hline
		\end{tabular}
	\end{center}
	$^1$ Our method on the best camera
	
	$^2$ Method of~\cite{debard_camera-based_2016} 
	on the best camera
	\label{table:final_results}	
	\vspace*{-2.5em}
\end{table}

\section{CONCLUSION} \label{sec:conclusion}

In this study, we brought a new perspective on fall detection solutions focused on the application in nursing homes. This vision has led to a new \gls{cnn} training strategy driven by a realistic alarm rate metric and a decision-making process that fits medical staff expectations. The presented solution has proven to detect 86.2\% of falls while producing only 11.6\% of false alarms in average on the considered databases. The analysis of false alarms has shown that in most cases they occur when the person sits down heavily, stands up after a fall or gets down to pick up something on the ground.

Our future works on that topic include the implementation of a spatial filter such as semantic background segmentation and an increase of the number and diversity of data in order to enhance the results. The system has been tuned and tested on fall videos simulated by performers, hence the next step would be to conduct a clinical study. Another opportunity would be to leverage on multiple cameras data fusion as in Multicam database within results analysis shows that a fall is always detected by at least one camera over all.

\addtolength{\textheight}{-12cm}   

\section*{ACKNOWLEDGMENT}

This work was founded by the European Union, the Britanny region in France and the French city of Rennes through the AAP FEDER - SilverConnect project.

The authors thank the "Pole Saint-Helier" which conducted the study in three different specialized homes for the elderly in the metropolitan area of Rennes, France.
\vspace*{-0.8em}
\bibliographystyle{IEEEbib}
\bibliography{bibliography}

\end{document}